\documentclass[letterpaper, 10 pt, conference]{ieeeconf}  

\IEEEoverridecommandlockouts                              

\usepackage{graphicx} 
\usepackage[utf8]{inputenc} 
\usepackage[T1]{fontenc}    
\usepackage{hyperref}       
\usepackage{url}            
\usepackage{booktabs}       
\usepackage{amsfonts}       
\usepackage{nicefrac}       
\usepackage{microtype}      
\usepackage{wrapfig}

\usepackage{mathtools}
\usepackage{amssymb}
\usepackage{amsmath}
\usepackage{xspace}
\usepackage{graphicx} 
\usepackage{makeidx}  
\usepackage{graphicx}
\usepackage[dvipsnames]{xcolor}
\usepackage{subcaption}
\usepackage{amssymb}
\usepackage{amsmath}
\usepackage{hhline}
\usepackage{textcomp}
\usepackage{thmtools}
\usepackage{thm-restate}
\usepackage{hyperref}
\usepackage[capitalise]{cleveref}
\crefname{appendix}{Appendix}{Appendices}
\Crefname{appendix}{Appendix}{Appendices}
\usepackage{multirow}
\usepackage{array}
\usepackage{calc}

\newtheorem{definition}{Definition}
\newtheorem{problem}{Problem}

\newtheorem{assumption}{Assumption}
\newtheorem{theorem}{Theorem}
\newtheorem{proposition}{Proposition}

\newtheorem{corollary}{Corollary}

\usepackage{tikz}
\usetikzlibrary{positioning,arrows.meta,fit,backgrounds,calc}
\definecolor{cKnow}{HTML}{2F6DB5}
\definecolor{cBel}{HTML}{D9822B}
\definecolor{cPlan}{HTML}{2E8B57}
\definecolor{cEnv}{HTML}{6B6B6B}
\definecolor{cOut}{HTML}{B03A48}
\newcommand{\ytop}{3.30}
\newcommand{\ybot}{0.05}
\tikzset{
  panel/.style={draw=#1!55, fill=#1!5, rounded corners=6pt, line width=0.8pt, inner sep=0pt},
  ptitle/.style={font=\footnotesize\bfseries, text=#1!75!black, anchor=north west, inner sep=4pt},
  box/.style={draw=#1!70!black, fill=white, rounded corners=3pt, line width=0.6pt, align=center, font=\scriptsize, inner sep=2.5pt},
  chip/.style={draw=#1!65!black, fill=#1!18, rounded corners=2pt, line width=0.5pt, font=\scriptsize, minimum width=1.45cm, minimum height=0.40cm, inner sep=1pt},
  room/.style={circle, draw=cBel!75!black, line width=0.6pt, minimum size=0.60cm, inner sep=0pt, font=\tiny},
  obj/.style={rectangle, rounded corners=1pt, draw=cBel!75!black, line width=0.5pt, minimum width=0.66cm, minimum height=0.36cm, inner sep=0pt, font=\tiny},
  flow/.style={-{Stealth[length=2.2mm,width=1.8mm]}, line width=0.8pt, draw=black!65},
  thin/.style={-{Stealth[length=1.6mm,width=1.3mm]}, line width=0.5pt, draw=black!55},
  sloop/.style={-{Stealth[length=1.2mm,width=1.0mm]}, line width=0.45pt, draw=cBel!85!black},
}

\definecolor{combsblue}{rgb}{0.00,0.35,0.85}

\definecolor{combsred}{rgb}{0.85,0.00,0.00}

\definecolor{combspurple}{rgb}{0.55,0.00,0.65}

\usepackage[normalem]{ulem}

\newcommand{\cn}{\textsc{ConceptNet}}
\newcommand{\gcombs}{\gamma_{\textsc{combs}}}
\newcommand{\lgt}{\operatorname{logit}}
\newcommand{\clipm}{\textsc{clip}}

\title{\LARGE \bf COMBS: Concealed Object Multi-modal Bayesian Search}

\author{Taylor Bergeron$^{1}$ and Kevin Leahy$^{1}$
\thanks{$^{1}$ Department of Robotics Engineering, Worcester Polytechnic Institute, Worcester, MA 01609.
        {\tt\small tabergeron@wpi.edu}}%
}

\begin{document}
\maketitle
\thispagestyle{empty}
\pagestyle{empty}

\begin{abstract}

    This work addresses the problem of finding a concealed object in an environment by unifying formalisms for partially observable environments with multi-modal priors drawn from a mix of ontologies, semantic networks, LLMs, and VLMs. We propose a concealment-adaptive mixed-observability MDP (CA-MOMDP), a Bayes-adaptive setting where uncertainty is over both object locations and which nodes can conceal them, held as a hypothesis with a bounded ontology prior and resolved only by interaction. We define a hierarchical transition system (HTS) to abstract the environment into room and object nodes that can be planned over via our CA-MOMDP. We further propose a COMBS-score: a multi-modal fusion of a human informed semantic network, LLM, and VLM, to initialize the belief over object locations, which grounds our abstractions and reduces reliance on a single model. Belief is maintained via Bayes' rule, which never lets a look rule out a closed container. We prove a present target is found in a bounded number of searches under exact perception and that its location is never ruled out given missed detection under noisy perception. We ablate each component to demonstrate their significance. Across five simulator configurations, from a gridworld to Matterport3D with a real detector, COMBS finds 92--100\% of in-distribution and 92--100\% of out-of-distribution targets, matches or beats the baselines in processing speed, significantly outperforms a per-step LLM policy at lower cost on the larger scenes, recovers more out-of-distribution placements than a TidyBot ranking, and matches an LLM-only prior at a fraction of its per-decision latency. COMBS is the only standalone method that can declare a target absent at a stated error level, and the per-step LLM policy offered a termination action abandons 12--22\% of targets that were present, against 1.6--3.2\% for COMBS on the same episodes.
   
\end{abstract}

\section{Introduction}

Modern perception models give a robot many partial, uncertain views of a scene, but a planner needs a belief it can act on, and building that belief in a principled way remains an open problem. Building an abstraction of the environment to reason over requires grounding the perception observations, which can draw on learned models, ontologies, and knowledge of how the environment layout limits what the robot can see. Consider searching for a target object that may be fully occluded in a partially known environment. Solving this task requires principled fusion of these data streams together with a hierarchical search that accounts for perception and observation uncertainty.

Semantic scene structure has largely asked what structure best represents a scene for a downstream task regardless of exploration techniques, but the structures typically do not account for what perception has not seen, meaning it becomes difficult to reason about where an object may be--a requirement for this problem~\cite{armeni_3d_2019, maggio_bayesian_2025}.
Works on semantic search consider a single data source or pretrained model and only consider a fully observable environment, which are not robust to LLM-related drawbacks or out-of-distribution data ~\cite{gintingSEEKSemanticReasoning2024, shahNavigationLargeLanguage2023, zhouESCExplorationSoft2023, linLLMbasedAgentsSuffer, chaplotObjectGoalNavigation2020}. 
Some works focus on searching for partially to fully occluded objects using existing models
but require spatial descriptions of where an object can be found ~\cite{haSemanticAbstractionOpenWorld2022, gadre_cows_2023, merlin_robot_2024}.  
Other occluded-object search works train a network to guide container search, but 
explore each container in turn rather than trading off opening against searching elsewhere and
fail on out-of-distribution placements~\cite{kurenkovSemanticGeometricModeling2021}.
Belief-based search models uncertainty about the locations of things already known to exist, rather than uncertainty over what exists at all, which is required for searching for a concealed target~\cite{aydemir_active_2013, 8793888}. 
Manipulation-based approaches search for an occluded object by moving clutter~\cite{danielczuk2019mechanical}, 
sometimes reasoning about target-object geometry relative to the scene~\cite{7759839}, but they do not consider higher-level exploration. Conversely, household domain approaches~\cite{kantHousekeepTidyingVirtual, wu2023tidybot} focus on placement, so the answer is a destination to execute rather than a hypothesis to test, and a wrong guess is never contradicted by observation.

We develop a framework, Concealed Object Multi-modal Bayesian Search (COMBS) to both leverage environment hierarchy though abstraction and semantic knowledge through multi-modal data fusion to find an object without any additional input or location descriptions.

Our contributions are: 1) a hierarchical transition system (HTS) whose edges are the robot's actions and whose nodes grow only when an action reveals objects, the structure concealed-object search requires, since a hidden object is reached only by opening what holds it; 2) the CA-MOMDP,  a factored Bayes-adaptive mixed-observability formulation, planned over the HTS, in which the robot's uncertainty covers not only where the target is but which nodes could be hiding it; and 3) a multi-modal prior, fused in log-odds with bounded influence from any single source, which initializes the CA-MOMDP's belief.


\section{Background and Problem Formulation}

Consider a robot tasked with finding an object in a house, such as a fork, but without location information provided. Using common knowledge, the robot should search within cabinets, drawers, and on surfaces based on the semantic relevance to the desired object. We desire an approach that incorporates semantics and succeeds when the object is occluded (in a drawer) or visible (on a table) and in in-distribution (table) or out-of-distribution (bathroom) locations.

To ensure that our approach is non-myopic, we consider a partially observable Markov decision process (POMDP), defined as a tuple $(S, A, \Omega, O, T, R)$, where $S$ is the set of states (which cannot be directly observed); $A$ is a set of actions; $\Omega$ is the observation space; $O(\omega,s)$ is the observation function for the probability of observing $\omega$ in state $s$; $T(s'|s,a)$ is the transition function for the probability of transitioning to $s'$ from $s$ by taking action $a$; $R(s,a,s')$ is the reward function for transition from $s$ to $s'$ by taking action $a$.

We would like to better model the states of the environment based on their occlusion. A mixed observability MDP (MOMDP) is an extension of a POMDP, where rather than assuming the entire environment state is hidden and must be estimated, a MOMDP splits the environment into a visible component and a hidden component~\cite{ong2010planning}.

Bayes-adaptive construction applied to a POMDP yields a BA-POMDP~\cite{ross2011bayesian}. Bayes-adaptive construction handles a model value that is fixed but unknown by moving it into the hidden state with a prior, so that acting teaches the agent its own model\cite{ross2011bayesian}. Its belief updates are principled, but it inherits the POMDP's entirely hidden environment state and adds an unknown for every value of the transition and observation functions. When every value is an estimate, no observation is ever conclusive, so we cannot provide guarantees, rendering a BA-POMDP intractable for our problem.

We desire an out-of-distribution robust solution to robot exploration considering uncertainty by holding a distribution over the target's location on a node set consisting of rooms and objects in the environment, that grows as the robot reveals contents. 
The approach should efficiently search the environment, based on expected detection and search effort, balancing tradeoffs for exploring rooms and opening containers.
As exploration progresses and the target object is not observed, the updated belief should guide the search.

\begin{problem}
\label{problem}
 Given a floor plan 
 and a set of pretrained models as priors, create a zero-shot non-myopic object search that considers the chance an object is entirely absent from the environment or is entirely occluded and
 that can effectively handle both in-distribution and out-of-distribution object placement, with guarantees on performance.
 \end{problem}


\begin{figure*}
\centering
\resizebox{\textwidth}{!}{%
\begin{tikzpicture}
\node[room, fill=cKnow!10, draw=cKnow!70!black] (fK) at (0.50,2.40) {Kit.};
\node[room, fill=cKnow!10, draw=cKnow!70!black] (fL) at (1.50,2.40) {Liv.};
\node[room, fill=cKnow!10, draw=cKnow!70!black] (fB) at (1.00,1.60) {Bath};
\draw[line width=0.6pt, cKnow!70!black] (fK)--(fL) (fL)--(fB) (fK)--(fB);
\node[font=\scriptsize, align=center] (fpl) at (1.00,0.95) {floor plan};
\node[chip=cKnow, minimum width=1.7cm] (query) at (1.00,0.40) {target $l_G$: \textit{fork}};
\begin{scope}[on background layer]
  \node[panel=cKnow, fit={(0.0,\ytop) (2.0,\ybot)}] (P1) {};
\end{scope}
\node[ptitle=cKnow] at (P1.north west) {Inputs};

\node[chip=cKnow] (clip) at (3.40,2.35) {CLIP};
\node[chip=cKnow] (cn)   at (3.40,1.85) {ConceptNet};
\node[chip=cKnow] (llm)  at (3.40,1.35) {Llama 3};
\node[box=cKnow, text width=1.45cm, minimum height=1.25cm] (pool) at (5.35,1.85) {log-odds pool\\[2pt]$\gcombs(j)$};
\foreach \s in {clip,cn,llm} \draw[thin] (\s.east) -- (pool.west |- \s.east);
\node[chip=cKnow] (wn) at (3.40,0.50) {WordNet};
\node[box=cKnow, text width=1.45cm] (ont) at (5.35,0.50) {concealment\\prior $p_v$};
\draw[thin] (wn) -- (ont);
\begin{scope}[on background layer]
  \node[panel=cKnow, fit={(2.55,\ytop) (6.30,\ybot)}] (P2) {};
\end{scope}
\node[ptitle=cKnow] at (P2.north west) {Zero-shot priors};

\node[room, fill=cBel!60] (K) at (8.00,2.10) {Kit.};
\node[room, fill=cBel!25] (L) at (9.25,2.10) {Liv.};
\node[room, fill=cBel!8]  (B) at (10.50,2.10) {Bath};
\draw[line width=0.6pt, cBel!75!black] (K)--(L)--(B);
\node[obj, fill=cBel!45] (dr) at (7.30,1.25) {drawer};
\node[obj, fill=cBel!20] (fr) at (8.05,1.25) {fridge};
\node[obj, fill=cBel!8]  (ct) at (8.80,1.25) {counter};
\draw[line width=0.5pt, cBel!75!black] (K)--(dr) (K)--(fr) (K)--(ct);
\node[obj, dashed, draw=black!45, text=black!55, fill=white] (hid) at (7.30,0.62) {?};
\draw[line width=0.5pt, dashed, black!45] (dr)--(hid);
\draw[sloop] (K.115) to[out=115,in=65,looseness=5] (K.65);
\draw[sloop] (B.115) to[out=115,in=65,looseness=5] (B.65);
\node[font=\tiny, align=left, anchor=north west, inner sep=0pt] (leg) at (9.35,1.50) {shade: belief $b(v)$\\[1pt]dashed: not yet revealed\\[1pt]loop: self-loop $S^{explore}$};
\node[font=\scriptsize, anchor=south] (bel) at (9.55,\ybot+0.08) {belief $b$ and concealment $p_v$};
\begin{scope}[on background layer]
  \node[panel=cBel, fit={(6.85,\ytop) (11.20,\ybot)}] (P3) {};
\end{scope}
\node[ptitle=cBel] at (P3.north west) {CA-MOMDP belief on HTS};

\node[box=cPlan, text width=2.85cm, minimum height=1.25cm, font=\scriptsize] (pol) at (13.30,1.85)
  {$\arg\max_{(v,a)}=$\\[2pt]$\dfrac{\sum_{k\in\Phi(v,a)} b(k)\,\alpha\,q_a(k)}{\mathrm{cost}(u_t,v,a)}$};
\node[chip=cPlan, minimum width=1.10cm] (vis) at (12.65,0.62) {$S^{visit}$};
\node[chip=cPlan, minimum width=1.10cm] (exp) at (13.95,0.62) {$S^{explore}$};
\draw[thin] (pol.south -| vis) -- (vis);
\draw[thin] (pol.south -| exp) -- (exp);
\node[font=\tiny, anchor=north] at (vis.south) {move + look};
\node[font=\tiny, anchor=north] at (exp.south) {open / sweep};
\begin{scope}[on background layer]
  \node[panel=cPlan, fit={(11.75,\ytop) (14.85,\ybot)}] (P4) {};
\end{scope}
\node[ptitle=cPlan] at (P4.north west) {Search policy};

\node[box=cEnv, text width=1.95cm, minimum height=1.25cm] (env) at (16.50,1.85) {robot or\\simulator\\[1pt]{\tiny Gridworld,\\ \mbox{ProcTHOR}, MP3D}};
\node[box=cEnv, text width=1.95cm] (obs) at (16.50,0.55) {detection $\omega$\\revealed $\mathcal{C}(v)$};
\draw[thin] (env) -- (obs);
\begin{scope}[on background layer]
  \node[panel=cEnv, fit={(15.40,\ytop) (17.60,\ybot)}] (P5) {};
\end{scope}
\node[ptitle=cEnv] at (P5.north west) {Environment};

\node[box=cOut, fill=cOut!8, text width=0.95cm] (found)  at (18.95,2.30) {target found};
\node[box=cOut, fill=cOut!8, text width=0.95cm] (absent) at (18.95,0.95) {report absent};
\draw[flow, draw=cOut] (P5.east |- found) -- node[above, font=\tiny] {$\omega{=}1$}(found.west); 
\draw[flow, draw=cOut] (P5.east |- absent) -- node[above, font=\tiny] {$L_t{<}\ell$}(absent.west); 

\draw[flow] (P1.east |- 0,1.85) -- (P2.west |- 0,1.85);
\draw[flow] (pool.east) -- node[above, font=\scriptsize]{$b_0$} (P3.west |- pool.east);
\draw[flow] (ont.east)  -- node[above, font=\scriptsize]{$p_v$} (P3.west |- ont.east);
\draw[flow] (P3.east |- pol.west) -- (pol.west);
\draw[flow] (pol.east) -- (env.west);

\draw[flow, draw=cBel!85!black] (P5.south) -- ++(0,-0.35) -| (P3.south)
  node[pos=0.25, below=1pt, font=\scriptsize, text=black]
  {update belief on the searched nodes, expand HTS, split previous belief over revealed children};
\end{tikzpicture}}
\caption{COMBS overview. Zero-shot priors initialize a belief over a hierarchical transition system that grows as the robot reveals contents. The policy selects the search with the highest expected detection per unit cost, and each observation that does not reveal the target updates the belief until the target is found or the absence test fires.}
\label{fig:overview}
\end{figure*}
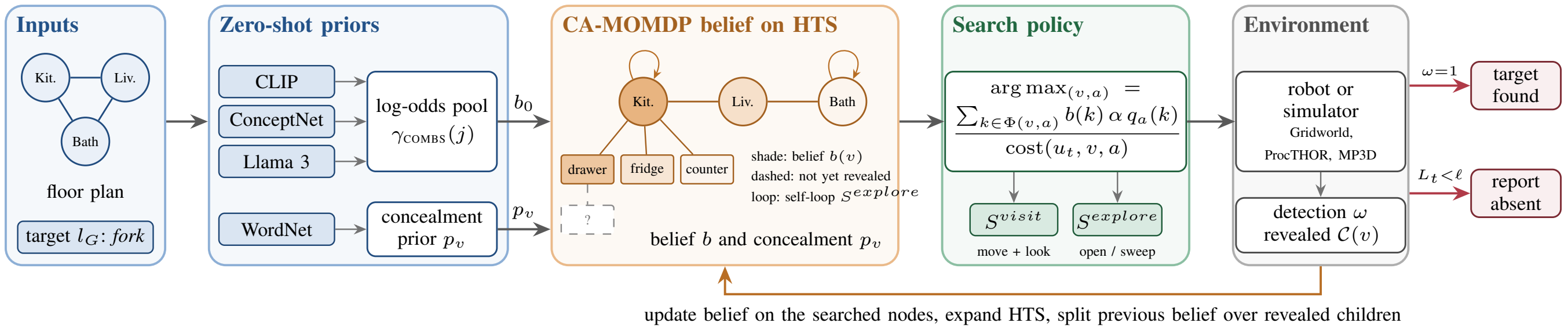

\section{Methodology}
\label{sec:method}

To solve Problem~\ref{problem} propose the COMBS pipeline, shown in Figure~\ref{fig:overview}. We propose a hierarchal transition system (HTS) to provide a search structure with edges as actions with cost and states that represent contents not yet observed, so a belief over concealed objects can be planned, updated, and proven complete directly on the HTS (Def.~\ref{def:hts}). The HTS is initialized with rooms and their corresponding belief from zero-shot priors. We cast the search as a concealment-adaptive mixed-observability MDP (CA-MOMDP) to reason over the chance an object is present or absent and visible or occuluded (Sec.~\ref{sec:camomdp}). The robot follows the search policy and 
updates existing nodes' values or adds new nodes based on its observations
until the target is found or reported absent.


\subsection{Environment abstraction}
\label{sec:envmodel}

To perform search we require a structure that can represent the environment concisely, the way an agent can interact with the environment, and the cost associated with those actions, while maintaining belief over the environment understanding.

\begin{definition}[\textbf{Scene}]
\label{def:scene}
A scene is a finite set of rooms, joined by the doorways of a given floor plan, and a finite set of objects. Each object is an instance of a semantic class drawn from a finite class set $AP$ and is located in exactly one room or inside exactly one other object. A class may have any number of instances. An object that holds objects is a container, and a container may hide its contents from view, in which case they are unknown until it is opened. Containment nests so the contents of a container may themselves be another container.
\end{definition}

We plan over a scene to find a target of semantic class $l_G \in AP$ such that cost is minimized. 

Since a container may hide its contents, no structure built prior to the run can list every state; states must instead appear as the agent acts on the scene. 
Since the robot's actions are physical, (e.g., crossing a doorway, opening a container, or examining a place without leaving it) each action carries a cost. Relational graphs such as a scene graphs do not natively capture such costs. We therefore consider a weighted transition system, which is a tuple of states, actions, a transition relation, a weight function assigning each transition a positive length, initial states, atomic propositions, and a labeling function $L$ recording which $l \in AP$ hold at each state~\cite{baier2008}. A weighted transition system has no containment among states, so we extend it into the hierarchical transition system.

\begin{definition}[\textbf{Hierarchical transition system (HTS)}]
\label{def:hts}
The HTS of a scene is a weighted transition system in which the states $\mathcal{V}$ are the scene's rooms together with one state per \textit{room} class per location, the state standing for all instances of that class located in that room or container; the labeling function $L$ assigns each state the singleton holding its class, written $L(v) = \{l_v\} \subseteq AP$; and layer $\mathcal{V}^k\subseteq \mathcal{V}$ holds the states at nesting depth $k$, the rooms being $\mathcal{V}^0\subseteq\mathcal{V}$.
The transition relation is the union of $E_\subset:\mathcal{V}^k\times\mathcal{V}^{k+1}$, one containment transition from each state to each state located in it; $E_\leftrightarrow \subseteq \mathcal{V}^0 \times \mathcal{V}^0$, one adjacency transition per doorway; and a self-loop at every state, the transition that acts at a state without leaving it. The actions are the choices of transition to execute, and the initial state is the robot's start room. 
The robot plans over the revealed part $\mathcal{V}_t \subseteq \mathcal{V}$, which holds $\mathcal{V}^0$ and the parent of each of its members and is not known in advance: executing a transition at $v$ at step $t$ may extend $\mathcal{V}_t$ by any subset of $v$'s children, so $\mathcal{V}_t \subseteq \mathcal{V}_{t+1}$.
\end{definition}

We define $d_t(u,v)$ for shortest-path distance in the HTS over $\mathcal{V}_t$, with edge lengths from the map and a fixed positive length for containment transitions.

\subsection{Multi-modal Prior}
\label{sec:multimodal}

The prior must not depend on any one model, since each of the available sources fails differently and none is trained on where household objects are kept. We combine three sources, CLIP, ConceptNet, and Llama, to create a \textbf{multi-modal prior}, so that a failed or uninformative one is absorbed by the other two, with \cref{prop:bounded} bounding the influence of any source.

\paragraph{CLIP} is a VLM trained on (image, text) pairings.  We use CLIP because it embeds any label, so it supplies a similarity for every pair of node labels. 

We assume the cosine similarity between two CLIP vectors is proportional to location plausibility.

We take the cosine similarity between the embedding of the goal label $l_G \in AP$ and each node label $l \in AP$. For some labels, this will be the only non-default non-zero value.

\Cref{eq:pool} combines probabilities, so each source must supply one for every candidate node, which CLIP inherently does not. Since cosine is positive, we divide each cosine by the total,
\begin{equation}
p^{\clipm}_j = \frac{\cos(\ell_G, \ell_j)}{\sum_{k \in \mathcal{V}_0} \cos(\ell_G, \ell_k)}.
\label{eq:clipnorm}
\end{equation}

\paragraph{ConceptNet} ConceptNet~\cite{speer_conceptnet_2017} is a semantic network from the crowd-sourcing project Open Mind Common Sense. Its data originates from crowd-sourced resources, expert-created resources, and games with a purpose. This is our only human informed dataset. 
For ConceptNet, we assume that
assertion-confidence mass is a usable proxy for location plausibility, and the \texttt{AtLocation} list is a partial rather than exhaustive set of plausible locations.

We normalize with mass held in reserve for locations the query did not return:
\begin{equation}
\begin{split}
p^{\cn}_j = \frac{w_j + a}{\sum_{k=1}^{m} w_k + a(m+1)},
\\
p^{\cn}_\varnothing = \frac{a}{\sum_{k=1}^{m} w_k + a(m+1)},
\label{eq:cnsmooth}
\end{split}
\end{equation}
where $p^{\cn}_\varnothing$ is the reserve. We set $a = \min_k w_k$, the weakest weight returned for that query. Each query's \texttt{AtLocation} weights may differ, so a fixed $a$ is unsuitable.
A label absent from the query result is assigned the reserve mass $p^{\cn}_\varnothing$ itself. 

\paragraph{Llama} We use Llama 3 8B parameter pretrained variant, an auto-regressive language model that uses an optimized transformer architecture, trained on publicly available online text data. We use Llama because it is another way to embed common sense reasoning into our method.
To interpret the data, we provide Llama the query "in a house I can find $l_G$ in the" and take softmax over the top $k=10$ logits. Additional prompts and $k$ values were tested for semantic relevancy but the survey is omitted for space.

\subsubsection{Combining Data Sources}

Log-odds is typically used in system consensus and in ML to combine predictions from logistic regression models.
The three sources normalize over different index sets, so we project each onto $\mathcal{V}$ before pooling. Mass on labels outside $\mathcal{V}$ is discarded, nodes the source does not report take its reserve mass, and the result is renormalized
\begin{equation}
\tilde{p}^{\,s}_j = \frac{\hat{p}^{\,s}_j}{\sum_{k \in \mathcal{V}_0} \hat{p}^{\,s}_k},
\qquad
\hat{p}^{\,s}_j = \begin{cases} p^s_j & j \text{ reported by } s,\\ p^s_\varnothing & \text{otherwise.}\end{cases}
\label{eq:project}
\end{equation}
so that~\eqref{eq:pool} compares node by node over one index set.

For each node we pool the three sources' opinions on whether the target is there, working in odds space so that the contributions add
\begin{equation}
\gcombs(j) \;=\; \sigma\!\left(\sum_{s} \omega_s \, \lgt\bigl(\tilde{p}^{\,s}_j\bigr)\right),
\qquad \sum_s \omega_s = 1,\; \omega_s > 0,
\label{eq:pool}
\end{equation}
with $\sigma$ the logistic function and $\omega_s = 1/3$ unweighted. Each $\lgt(\tilde{p}^{\,s}_j)$ is the log-odds that source $s$ places on the target being at $j$, and summing them with weights that sum to one is the accumulation of evidence in odds form. The prior is obtained by normalizing $\gcombs$ over the nodes and flooring it,
\begin{equation}
b_0(j) = (1-\varepsilon)\,\frac{\gcombs(j)}{\sum_{k \in \mathcal{V}_0} \gcombs(k)} + \frac{\varepsilon}{|\mathcal{V}_0|},
\label{eq:prior}
\end{equation}
for $\varepsilon \ge 0$. 
Every source is strictly positive on every node after~\eqref{eq:cnsmooth} and~\eqref{eq:project}, so the pool cannot produce a zero belief it cannot recover from, as~\eqref{eq:update} and~\eqref{eq:split} only rescale a node's belief.

\begin{proposition}[\textbf{Bounded influence of a source}]
\label{prop:bounded}
If every source assigns node $j$ strictly positive mass, then $\gcombs(j) > 0$. If source $s$ reports its reserve $\tilde{p}^{\,s}_\varnothing$ in place of a value $\tilde{p}^{\,s}_j > \tilde{p}^{\,s}_\varnothing$, the log-odds of $\gcombs(j)$ fall by exactly $\omega_s\bigl[\lgt(\tilde{p}^{\,s}_j) - \lgt(\tilde{p}^{\,s}_\varnothing)\bigr]$, a finite quantity fixed by~\eqref{eq:cnsmooth}.
\end{proposition}

\subsection{Ontology prior}

Deciding whether to open a node before ever touching it needs a class-level judgment about whether it can contain other objects, which is a question for a formal ontology rather than for the location sources above. An incorrect ontology prior costs unnecessary search actions, since no prior can rule a location out of the belief.

We query WordNet~\cite{miller1995wordnet}, whose coverage of household nouns is essentially complete and whose senses are what open-vocabulary detectors produce as labels. A label is containment-capable iff one of its noun senses has a hypernym path to \texttt{container.n.01}, \texttt{appliance.n.02}, or \texttt{fixture.n.01}; a short curated list corrects labels WordNet misfiles (e.g.\ cabinet, filed under furniture). Since classification is binary, it must be balanced against missing entries, by adding two phantom votes to WordNet's real vote $k \in \{0, 1\}$,
\begin{equation}
p_v \;=\; \frac{k + 1}{3},
\qquad
p_v = \tfrac{1}{2} \;\text{ for a label WordNet cannot map},
\label{eq:ontprior}
\end{equation}
so a containment-capable label is bounded in $[\nicefrac{1}{3}, \nicefrac{2}{3}]$. Only observation moves $p_v$ outside that band.

\subsection{Detection rate}
When a robot searches a physical world, it can miss a present target by another object or a poor camera pose can hide it from view, or the detector mislabels the target or the container holding it. We represent these with a single detection rate $\alpha$
, the probability that a search detects a target it covers, so a missed target's container keeps belief and can be searched again rather than being set to zero.

\subsection{The CA-MOMDP}
\label{sec:camomdp}

The search needs a decision model that reasons over a prior belief about where the target is and accounts for the agent's own actions changing what is occluded. 

\begin{definition}[\textbf{CA-MOMDP}]
\label{def:camomdp}
A concealment-adaptive MOMDP is a tuple $(\mathcal{S}, \mathcal{A}, \Omega, O, \rho, r, b_0)$. A state $s = (u, \mathcal{V}_t, X, c) \in \mathcal{S}$ records the robot's node $u$, the nodes $\mathcal{V}_t$ revealed so far, the target location $X$, and a static concealment assignment $c$ unknown to the agent, where $c_v = 1$ iff node $v$ hides its contents from view. Of these, $(u, \mathcal{V}_t)$ is fully observable and $(X, c)$ is not. The actions $\mathcal{A}$ are the transitions of the HTS, where $S^{visit}$ takes an adjacency or containment transition and examines what arrival exposes, and $S^{explore}$ takes the self-loop, examining the node itself and revealing its contents. The transition function $\rho$ is deterministic on $(u, \mathcal{V}_t)$, an observation $\omega \in \Omega = \{0,1\}$ records whether the target was detected and is drawn from $O$ of~\eqref{eq:obsfn}, and $r$ is the transition cost the environment reports. The initial belief $b_0$ pairs the prior over $X$ of~\eqref{eq:prior} with the ontology prior over $c$ of~\eqref{eq:ontprior}.
\end{definition}

What the BA-POMDP is to the POMDP, the CA-MOMDP is to the MOMDP. The Bayes-adaptive construction applied to its hidden state, narrowed to one binary value per node representing wether the node conceals its contents. The target's is an ordinary hidden state, which the MOMDP already carries.

Since only $(X, c)$ is hidden the belief can be captured by the distribution $b$ over the target's location and one Bernoulli $p_v$ per node. The belief is updated according to~\eqref{eq:update} and~\eqref{eq:split}. The concealment prior depends only on what kind of object $v$ is, so an ontology can supply it instead of data and the search remains zero-shot.

Updating $b$ and $p_v$ requires knowing which nodes an action reaches and how reliably it detects a target there. The sensor scope $\Phi(u,a) \subseteq \mathcal{V}$ is the set of nodes that action $a$ reaches at node $u$; a search \emph{examines} exactly the nodes in its scope. For $S^{visit}$ it is the children of $u$ already in $\mathcal{V}_t$, together with $\{u\}$ once $u$ has been explored. For $S^{explore}$ it is $\{u\} \cup \mathcal{C}(u)$, the node itself and any contents it reveals. The same scopes apply at every node, so~\eqref{eq:update} and~\eqref{eq:policy} are node type agnostic. Either action may also reveal children, and~\eqref{eq:split} applies whenever the child set grows. A constant detection rate $\alpha \in (0,1]$ per simulation is the chance that a search detects a target it covers, with $\alpha = 1$ where perception is exact.

\begin{definition}[Occlusion-aware observation function]
For $\omega \in \{0,1\}$,
\begin{equation}
O(\omega = 1 \mid X = j, u, a) =
\begin{cases}
\alpha, & a \text{ is the self-loop at } j,\\
\alpha\,(1 - c_j), & a \text{ looks at } j,\\
0, & j \notin \Phi(u,a),
\end{cases}
\label{eq:obsfn}
\end{equation}
\end{definition}
The detection probability of~\eqref{eq:obsfn} has two factors. $\alpha$ is the chance perception recognizes a target it can see. $q_a(j)$ is the chance the target at $j$ is visible after action $a$. The self-loop opens $j$, so everything inside is visible and $q_a(j) = 1$. A look sees only the outside, so the target is visible only if $j$ does not conceal it, $q_a(j) = 1 - c_j$. An action that does not reach $j$ sees nothing, $q_a(j) = 0$. The agent does not know $c_j$, so for a look it plans with the expected value $1 - p_j$.
\begin{assumption}
No false positives. $O(\omega = 1 \mid X, u, a) = 0$ whenever $X \notin \Phi(u,a)$, so a detection ends the search.
\end{assumption}

When a node is revealed, the robot cannot tell whether it conceals its contents. An agent must be able to distinguish a drawer that has been searched with out identifying the target from a closed drawer whose exterior has been examined. Otherwise, the agent would never choose to open a closed drawer. The agent instead holds $p_v = \Pr[c_v = 1]$, initialized from~\eqref{eq:ontprior}, so a look at $v$ detects with probability $(1-p_v)\,\alpha$ and never determines $c_v$. The self-loop of $S^{explore}$ determines $c_v$ as it examines the interior.

\subsection{Search Algorithm}\label{sec:algorithm}

The policy searches the pair $(v,a)$ maximizing expected detection per unit cost,
\begin{equation}
a_t = \arg\max_{(v,a)\,:\, v \in \mathcal{V}_t} \frac{\sum_{k \in \Phi(v,a)} b_t(k)\,\alpha\,q_a(k)}{\mathrm{cost}(u_t, v, a)},
\label{eq:policy}
\end{equation}
The cost includes the travel $d_t(u_t,v)$ to reach $v$, and the action's own cost $c_a > 0$, with opening costing more than looking. The travel term stops the robot bouncing between distant rooms of near-equal belief. 
The action term ensures that explore is not always selected over visit. $\Phi(v, S^{explore})$ contains $\Phi(v, S^{visit})$, so exploring never scores a smaller numerator and would get chosen as the action at every node if the two cost the same. Costs are bounded, $0 < c_{\min} \le \mathrm{cost} \le c_{\max}$. The robot travels directly to the selected node and acts there, and it neither observes the rooms it passes through nor re-plans until that action is done. A concealed location is not yet a node of $\mathcal{V}_t$ and is reached only by opening its parent; this is the explore-versus-search tradeoff.

\subsection{Belief on an HTS that grows}

Belief cannot be held over objects not yet seen, so the support of $b$ must expand as contents are revealed, without discarding what searches not finding $l_G$ have already established. We therefore read a node's belief as
\begin{equation}
\begin{split}
    b(v) \;=\; \Pr\bigl[\,X \text{ lies in the region of } v \text{ but} \\ \text{  not in any revealed descendant of } v\,\bigr].
\label{eq:nodemeaning}
\end{split}
\end{equation}
Before anything at $v$ is revealed there are no descendants to exclude, so $b(v)$ is the probability that $X$ lies anywhere in $v$'s region. Each revealed child removes its own region from that event, until $b(v)$ covers only what $v$ exposes directly.

When an action at $v$ reveals children $\mathcal{C}(v)$, the belief at $v$ is subdivided among $v$ and its new children in proportion to~\eqref{eq:pool} applied to that sibling set,
\begin{equation}
\begin{split}
b'(c) = b(v)\,\pi_v(c) \;\; \text{for } c \in \mathcal{C}(v) \cup \{v\},
\\
\pi_v(c) = \frac{\gcombs(c)}{\sum_{c' \in \mathcal{C}(v) \cup \{v\}} \gcombs(c')},
\label{eq:split}
\end{split}
\end{equation}
with every other node left unchanged and total mass preserved. \eqref{eq:split} is a conditioning operation, so no evidence is counted twice, and it touches only mass that no observation has yet conditioned, so the pool is never applied on top of a posterior.

\subsection{Belief update on a negative observation}

A real robot may mislabel, miss an object behind another, or fail to physically open a container. We design the update to account for perception errors, so the update never treats an observation that did not reveal the target (negative observation) as certain, so the update returns mass to everywhere the search did not reach. The one exception is exact perception, $\alpha = 1$, where examining a node rules it out, but visiting never does since a node may conceal the target.

\begin{definition}[Belief update on a negative observation]
Let $Z = 1 - \sum_{k \in \Phi(u,a)} b(k)\alpha\,q_a(k)$ be the probability that this search returns no $l_G$ given the target is in the scene. On observing $\omega = 0$,
\begin{equation}
b'(j) = \begin{cases}
b(j)\bigl(1 - \alpha\,q_a(j)\bigr)/Z, & j \in \Phi(u,a),\\[2pt]
b(j)/Z, & j \notin \Phi(u,a).
\end{cases}
\label{eq:update}
\end{equation}
\end{definition}
Every node outside $\Phi(u,a)$ strictly gains mass on a negative observation. This formulation is how the agent recovers from an out-of-distribution placement. For example, no prior ever endorses the bathroom, but repeated negative observation at other nodes raises the bathroom. A negative observation updates the current node's concealment hypothesis $p_v$ by the same Bayes' rule. If $v$ is concealed, a look can never find the target inside it, so a negative observation is evidence for concealment: $p'_v = p_v A_v/Z$, where $A_v$ is $Z$ recomputed without $v$'s own term, which rises with every failure. Opening $v$ shows its contents, so $c_v$ becomes known and needs no update.

\begin{proposition}[Support preservation]
\label{prop:support}
Under~\eqref{eq:update}, a positive belief becomes zero only when $\alpha = 1$ and the search is the self-loop at that node.
\end{proposition}
\begin{proof}
$Z > 0$ because the observation occurred, and $1-\alpha\,q_a(j) = 0$ only if $\alpha = 1$ and $q_a(j) = 1$, which is the self-loop, since a look has $q_a(j) = 1-p_j < 1$.
\end{proof}

\subsection{Terminating when the target is absent}

The robot must end the search when it determines there is no target in the scene, but no quantity in the belief can signal that. Each $Z_i$ is the probability that observation $i$ is negative given a present target, and an absent target makes every observation negative with certainty, so
\begin{equation}
L_t \;=\; \prod_{i \le t} Z_i
\;=\; \frac{\Pr[\,\text{all negative observations} \mid \text{target present}\,]}{\Pr[\,\text{all negative observations} \mid \text{target absent}\,]}
\label{eq:evidence}
\end{equation}
is a likelihood ratio between presence and absence. Once $L_t$ falls below a stated level $\ell$, the robot searches each unexamined room and each node the ontology prior deems capable of concealment. Absence is only reported when that node set search is complete, so absence is never declared while a likely container remains unopened.

\subsection{Completeness}\label{sec:completeness}

\begin{theorem}[Completeness]
\label{thm:complete}
Assume finitely many nodes, a stationary target, explores that reveal all children, and $\alpha = 1$. Then~\eqref{eq:policy} with~\eqref{eq:update} finds the target within a number of searches bounded by $|\mathcal{V}|$ and $c_{\max}/c_{\min}$.
\end{theorem}

\begin{proof}The deepest revealed node holding the target keeps positive belief until the target is found (\cref{prop:bounded,prop:support}), so its self-loop stays selectable. A self-loop at a node with positive belief zeroes it, at most $|\mathcal{V}|$ times. Any other selection must outscore such a self-loop, so it raises some $p_k \le 1$ by a factor of at least $1/(1-\eta)$, $\eta = c_{\min}/(|\mathcal{V}|^2 c_{\max})$ (\cref{eq:update}), which happens finitely often since $p_k \ge \nicefrac{1}{3}$.\end{proof} With $\alpha < 1$ no belief reaches zero (\cref{prop:support}), so a missed target stays recoverable.

\begin{corollary}[Completeness with unknown concealment]
\label{cor:concealment}
Let the conditions of \cref{thm:complete} hold with concealment unknown and $p_v$ initialized by~\eqref{eq:ontprior}. Then a present target is found within the bound of \cref{thm:complete}. With $\alpha = 1$, a node that conceals the target but starts at $p_v = 0$ is zeroed by its first look, and the target is never found.
\end{corollary}

\begin{proof}The first claim is \cref{thm:complete}, whose proof already uses $p_v$. With $\alpha = 1$ and $p_v = 0$, a look at a concealing node counts as a full examination, so it zeroes that node's belief and the node is never searched again.\end{proof}

\begin{corollary}[Completeness under target absence]
\label{cor:stopping}
Under \cref{thm:complete}, with searches conditionally independent given $X$ and $c$, the absence test and its cleanup find a present target with probability at least $1-\ell$ and report an absent target absent after finitely many searches. With $\alpha < 1$ this holds when $\alpha$ does not exceed the true detection rate.
\end{corollary}


\section{Experiment}

\subsection{Simulation}
To demonstrate the efficacy of COMBS, we implement our approach in semantically sound simulation environments.
We evaluate on three simulators, a Gridworld, ProcTHOR~\cite{deitkeProcTHORLargeScaleEmbodied2022}, and Matterport3D (MP3D)~\cite{chang2017matterport3d}, with ten scenes each and the five simulator configurations of \cref{tab:eval_sims}. Gridworld and ProcTHOR walk meet the conditions of \cref{thm:complete}. ProcTHOR camera and MP3D camera reveal only what a camera sees, and MP3D Detic also takes its labels from a detector, so each step relaxes one condition. The subroutines for $S^{visit}$ and $S^{explore}$ vary per simulator.

\subsubsection{Gridworld}
Each floor of an MP3D house becomes a single-floor grid scene with its annotated rooms, furniture, and labels. Actions are abstract, so $S^{visit}$ reveals a sample of a node's contents and $S^{explore}$ reveals all of them. Travel cost is grid distance between node's assigned cells.

 \subsubsection{ProcTHOR}
Houses come from ProcTHOR's training split and are kept if they have at least two connected rooms and a kitchen with an openable drawer or cabinet. $S^{visit}$ drives to a room's center or approaches an object, and $S^{explore}$ sweeps a room, views a non-container closely, or opens a container; this is the only simulator with 3 distinct subroutines for $S^{explore}$ Labels come from simulator metadata; ProcTHOR walk reveals every child on an explore, and ProcTHOR camera reveals only objects visible in the rendered frames.

\subsubsection{MP3D}
We use single-floor scans. The robot moves over each scan's panorama graph and reveals an object when rays from a panorama's camera reach it through the scan's mesh, and opening a container reveals all its contents. $S^{visit}$ goes and views the center of a room or approaches an object, $S^{explore}$ sweeps the room or but wereveals if an object has contents. MP3D camera uses the annotated labels. MP3D Detic labels the photographs with Detic~\cite{zhou2022detecting} over the LVIS vocabulary. A detection is assigned to the annotated object covering most of its mask, 
correct or not.

\begin{table}
\centering
\caption{Simulator configurations, as in \cref{tab:main}
}
\label{tab:eval_sims}
\setlength{\tabcolsep}{3pt}
\resizebox{\columnwidth}{!}{%
\begin{tabular}{@{}llllccl@{}}
\toprule
Simulation & Labels & Room explore & Container open & $\alpha$ & Rooms & Tests \\
\midrule
Gridworld       & exact & all objects & all contents & $1$    & 11.5 & no perception \\
ProcTHOR walk   & exact & all objects & all contents & $1$    & 5.4  & platform realism \\
ProcTHOR camera & exact & camera view & camera view  & $0.95$ & 5.4  & view-limited perception \\
MP3D camera     & exact & camera view & all contents & $0.95$ & 16.6 & real scanned homes \\
MP3D Detic      & Detic & camera view & all contents & $0.70$ & 16.6 & label noise \\
\bottomrule
\end{tabular}}
\end{table}

\subsection{Implementation}

The semantic sources are CLIP ViT-B/32, Llama~3~8B, and ConceptNet's public API, and WordNet~\cite{miller1995wordnet} supplies the concealment prior of~\eqref{eq:ontprior}. The queries are the sixteen objects every platform and placement source can express: apple, book, bowl, box, cup, fork, knife, laptop, mug, pan, pillow, plate, remote control, spatula, spoon, and vase. In-distribution placements follow SOMA's storage rules~\cite{bessler2021soma} and out-of-distribution placements follow Housekeep's implausible-placement votes~\cite{kantHousekeepTidyingVirtual}, neither of which the planner reads, and seven of every ten placements conceal the target. The planner never reads a simulator's openability or concealment data.

Calibrated $\alpha$ per simulator configuration is reported in Table~\ref{tab:eval_sims}. For camera simulator configurations, $\alpha$ is calibrated on held-out houses (five ProcTHOR validation houses and five MP3D scans) as the fraction of searches that find and report the target.
MP3D weights concealed and open target success $0.7$ and $0.3$ and counts a mislabeled container as a Detic miss. ProcTHOR camera uses the fraction of opened containers where the target is seen at first look. We set $\ell = 1/5$ and $\varepsilon = 0$.

\begin{table}
\centering
\caption{Knowledge sources. Robot columns are the priors the planner holds; world columns are the truths they estimate. \textsc{g}~=~gridworld, \textsc{p}~=~ProcTHOR, \textsc{m}~=~MP3D, $^\dagger$~=~evaluation.}
\label{tab:sources}
\setlength{\tabcolsep}{5pt}
\small
\begin{tabular}{@{}l cc ccc@{}}
\toprule
 & \multicolumn{2}{c}{Robot believes} & \multicolumn{3}{c}{World fixes} \\
\cmidrule(lr){2-3}\cmidrule(lr){4-6}
Source & $b_0$ & $p_v$ & Layout & $c_v$ & $X$ \\
\midrule
ConceptNet,CLIP,Llama     & \textsc{gpm} &             &             &             &             \\
WordNet                     &             & \textsc{gpm} &             &             &             \\
SOMA~\cite{bessler2021soma} &             &             &             & \textsc{gpm}  & \textsc{gpm} \\
Matterport                  &             &             & \textsc{gm}  &             &             \\
ProcTHOR assets             &             &             & \textsc{p}  & \textsc{p}  &             \\
Housekeep~\cite{kantHousekeepTidyingVirtual} & & & & & \textsc{gpm$^\dagger$} \\
\bottomrule
\end{tabular}
\end{table}

\subsection{Baselines}
Every method searches the same HTS with the same actions, costs, and episodes, but they differ in how they select the search order and what they consider for their decision. Tidybot~\cite{wu2023tidybot} queries the external LLM once per episode to rank the room and selects the top ranking without further reason. The LLM policy queries the external LLM at every step to select the best action among the revealed nodes, so while it can replan it holds no belief. COMBS variants replace the pooled COMBS-score for scoring locations. COMBS + LLM belief uses the external LLM, and COMBS + CLIP belief uses raw CLIP similarity; both keep WordNet's concealment prior and the noise-aware update, so they isolate the location prior.

\subsection{Results}

Throughout the results, $^{*}$ marks a significant paired difference from COMBS on the same episodes ($p<0.05$, McNemar or Wilcoxon).
Cost is the distance traveled plus the price of each sweep or opening, using grid distance for Gridworld, meters for ProcTHOR and MP3D, with an MP3D room sweep charged the length of its panorama tour.

\begin{table*}
\centering
\caption{Search performance on the simulators of \cref{tab:eval_sims}. Succ.: \% of present targets found, for in-distribution (ID) and out-of-distribution (OOD) placements. 10 houses × 2 placements (SOMA = ID, Housekeep = OOD) × 16 queries × 3 seeds.
}
\label{tab:main}
\footnotesize
\setlength{\tabcolsep}{3pt}
\resizebox{\textwidth}{!}{%
\begin{tabular}{@{}l cccc cccc cccc cccc cccc@{}}
\toprule
 & \multicolumn{4}{c}{Gridworld} & \multicolumn{4}{c}{ProcTHOR walk} & \multicolumn{4}{c}{ProcTHOR camera} & \multicolumn{4}{c}{MP3D camera} & \multicolumn{4}{c}{MP3D Detic} \\
\cmidrule(lr){2-5}\cmidrule(lr){6-9}\cmidrule(lr){10-13}\cmidrule(lr){14-17}\cmidrule(lr){18-21}
Method & ID & OOD & SPL & Cost & ID & OOD & SPL & Cost & ID & OOD & SPL & Cost & ID & OOD & SPL & Cost & ID & OOD & SPL & Cost \\
\midrule
COMBS & \textbf{99} & \textbf{100} & \textbf{0.46} & \textbf{120/119} & \textbf{100} & \textbf{100} & 0.46 & 133/133 & 92 & 92 & 0.46 & 136/124 & \textbf{99} & \textbf{100} & 0.53 & 147/146 & 94 & \textbf{98} & 0.52 & 173/146 \\
COMBS + LLM belief & 98 & \textbf{100} & $0.39^{*}$ & $154/151^{*}$ & \textbf{100} & \textbf{100} & $0.47^{*}$ & $110/110^{*}$ & 93 & 92 & 0.47 & $119/109^{*}$ & \textbf{99} & \textbf{100} & 0.56 & \textbf{142/142} & 94 & 97 & 0.54 & 169/153 \\
COMBS + CLIP belief & $87^{*}$ & $94^{*}$ & $0.39^{*}$ & $165/152^{*}$ & $71^{*}$ & $48^{*}$ & $0.37^{*}$ & $196/82^{*}$ & $71^{*}$ & $58^{*}$ & $0.41^{*}$ & $176/72^{*}$ & $96^{*}$ & \textbf{100} & $0.31^{*}$ & $262/259^{*}$ & $89^{*}$ & $93^{*}$ & $0.30^{*}$ & $283/242^{*}$ \\
TidyBot ranking & $95^{*}$ & $64^{*}$ & $0.35^{*}$ & $142/103^{*}$ & $96^{*}$ & $46^{*}$ & \textbf{0.49} & $\mathbf{95/70}^{*}$ & $89^{*}$ & $55^{*}$ & $\mathbf{0.49}^{*}$ & $\mathbf{97/74}^{*}$ & $97^{*}$ & $92^{*}$ & $\mathbf{0.64}^{*}$ & 157/143 & \textbf{95} & $94^{*}$ & $\mathbf{0.63}^{*}$ & $\mathbf{151/141}^{*}$ \\
LLM policy & $93^{*}$ & $89^{*}$ & $0.37^{*}$ & $303/251^{*}$ & \textbf{100} & \textbf{100} & 0.47 & $101/101^{*}$ & $\mathbf{97}^{*}$ & $\mathbf{94}^{*}$ & $0.46^{*}$ & $119/99^{*}$ & 98 & $94^{*}$ & $0.43^{*}$ & $233/202^{*}$ & \textbf{95} & $95^{*}$ & $0.42^{*}$ & $213/178^{*}$ \\
\bottomrule
\end{tabular}}
\end{table*}

\begin{table}
\centering
\caption{Ablations. Each cell is the change in success (percentage points) against COMBS on the same episodes. Run on 50\% of each simulator configuration.
}
\label{tab:ablation}
\footnotesize
\setlength{\tabcolsep}{3pt}
\resizebox{\columnwidth}{!}{%
\begin{tabular}{@{}l ccccc@{}}
\toprule
 & Grid- & ProcTHOR & ProcTHOR & MP3D & MP3D \\
Ablation & world & walk & camera & camera & Detic \\
\midrule
COMBS (success \%) & 99 & 100 & 94 & 99 & 94 \\
Uniform concealment prior & $-6^{*}$ & $-21^{*}$ & $-12^{*}$ & $-32^{*}$ & $-23^{*}$ \\
Concealment off & $-44^{*}$ & $-31^{*}$ & $-19^{*}$ & $-55^{*}$ & $-47^{*}$ \\
Uniform cost & $-3^{*}$ & +0 & +0 & +0 & -1 \\
Perception noise off ($\alpha = 1$) & n/a & n/a & +0 & +0 & -1 \\
\bottomrule
\end{tabular}}
\end{table}

\begin{figure}
\centering
\includegraphics[width=\columnwidth]{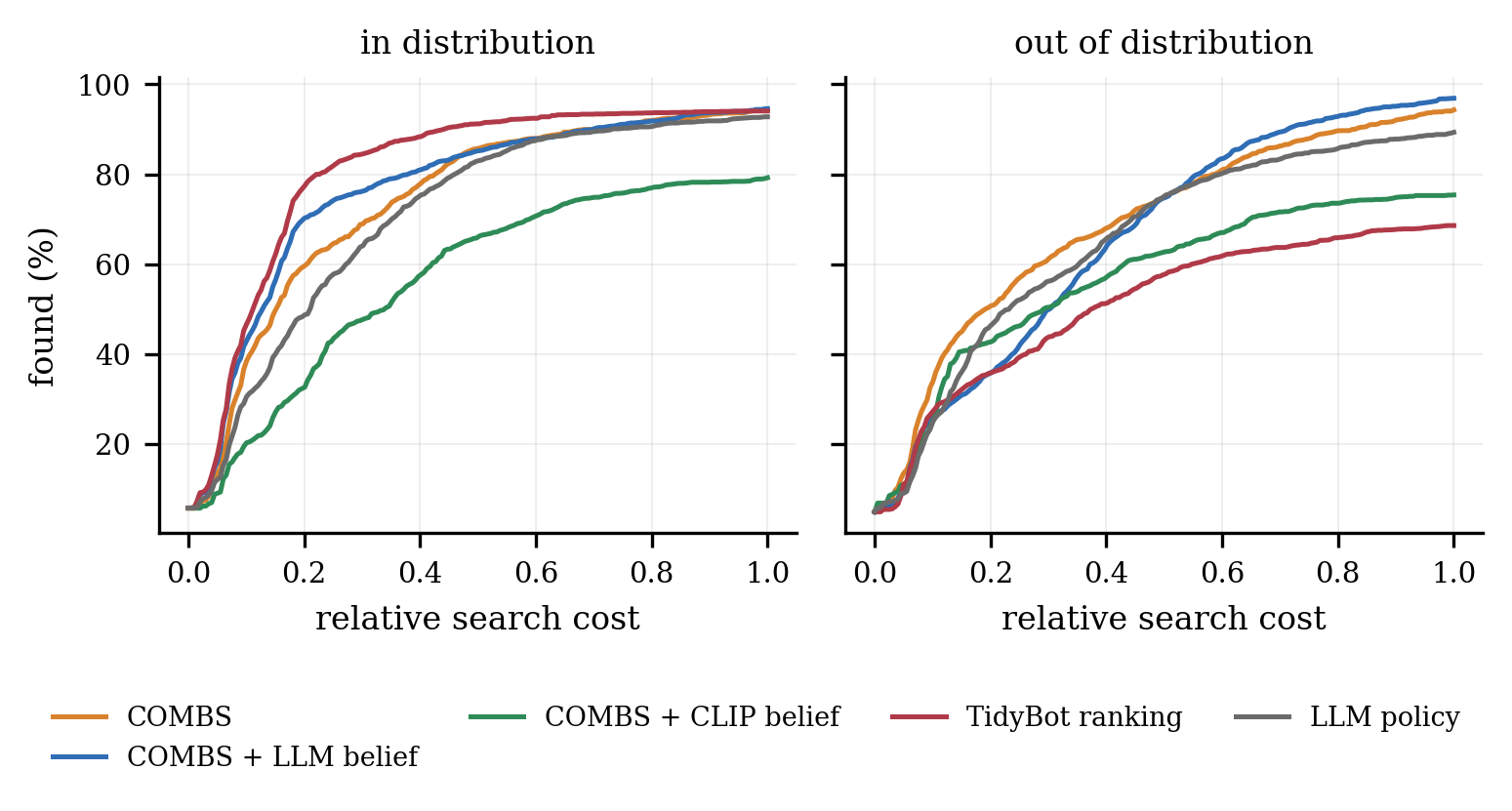}
\caption{Share of present targets found within a given search cost, for in-distribution and out-of-distribution placements, pooled over the five simulator configurations.
}
\label{fig:success_cost_grouped}
\end{figure}

\begin{table}
\centering
\caption{Absence reasoning at $\ell = 0.2$.
Present target episodes: 9 house--query pairs $\times$ 2 placements (SOMA = ID, Housekeep = OOD) $\times$ 3 seeds $\times$ 5 simulators. Absent target episodes: 32 house--query pairs $\times$ 1 seed $\times$ 5 simulators.
}
\label{tab:llm_absence}
\setlength{\tabcolsep}{3pt}
\resizebox{\columnwidth}{!}{%
\begin{tabular}{@{}l cc cc cc@{}}
\toprule
 & \multicolumn{2}{c}{Present ID} & \multicolumn{2}{c}{Present OOD} & \multicolumn{2}{c}{Absent target} \\
\cmidrule(lr){2-3}\cmidrule(lr){4-5}\cmidrule(lr){6-7}
Method & Found (\%) & \shortstack{Incorrectly\\declared (\%)} & Found (\%) & \shortstack{Incorrectly\\declared (\%)} & \shortstack{Correctly\\declared (\%)} & \shortstack{Steps to\\declare} \\
\midrule
COMBS                & \textbf{98.4} & \textbf{1.6} & \textbf{97} & \textbf{3.2} & 96 & 62.1 \\
COMBS + LLM belief   & 96.7 & 3.3 & \textbf{97} & \textbf{3.2} & \textbf{98} & 55.4 \\
LLM policy + absence & $85.2^{*}$ & $12.3^{*}$ & $74^{*}$ & $22.4^{*}$ & $88^{*}$ & \textbf{38.9} \\
LLM policy           & 97.5 & n/a & 96 & n/a & n/a & n/a \\
\bottomrule
\end{tabular}}
\end{table}

\Cref{tab:main} shows that COMBS finds nearly every present target on all five simulator configurations, in and out of distribution, and finds significantly more of them at lower cost than a belief built from CLIP similarity alone. With an LLM supplying the prior inside the same pipeline, COMBS + LLM belief matches COMBS on the MP3D configurations but search costs more on Gridworld, so the pooled prior holds up against a much larger model. The TidyBot ranking is sometimes cheaper on in-distribution targets but finds fewer out-of-distribution ones, and the per-step LLM policy, which drops the belief, costs more on the larger scenes
It overtakes COMBS in success only on the small ProcTHOR camera houses, so what COMBS adds beyond a competitive ranking is recovery from out-of-distribution placement and a principled way to stop. On the camera configurations COMBS never runs out of steps while searching. When it does not declare absence, COMBS finds every present target. When COMBS fails to find a target that is present, it is because COMBS wrongly declared the target absent and stopped, and a method that cannot declare absence can never lose an episode this way.

\Cref{fig:success_cost_grouped} shows how quickly each method finds targets as its accumulated cost, the distance traveled plus the price of each sweep and opening, grows. In distribution, the belief-prior baselines match COMBS. Out of distribution they separate: the TidyBot ranking flattens early, since a fixed ranking cannot revise a wrong guess; COMBS + CLIP belief keeps climbing but slowly, since its prior barely separates locations; the per-step LLM policy recovers but at 1.2--2.5 times COMBS's cost on Gridworld and MP3D. COMBS and COMBS + LLM belief climb to nearly every target, which is the recovery the negative-observation update provides.

\Cref{tab:ablation} shows that concealment reasoning is the component the results rest on: removing it, or replacing the ontology prior with a uniform one, causes the largest drops in success. The action cost appears independent from success. 

\Cref{tab:llm_absence} tests whether absence declarations can simply be prompted. The LLM policy, instructed it may stop whenever it believes the target is not present (recorded as LLM policy + absence), abandons roughly one in six present targets, twice as often out-of-distribution, while declaring absence on absent-target episodes less often than COMBS. The value of the absence machinery is not the ability to stop, which any prompt can grant, but ending the search only when no statistically credible location remains.

We measured that real detector labels are noisy, with a quarter of objects never labeled and half of first labels wrong and rarely corrected, so MP3D Detic tests the search under substantial label noise rather than a near-perfect detector, as our design intended.

\paragraph{Compute}

\begin{figure}
\centering
\includegraphics[width=\columnwidth]{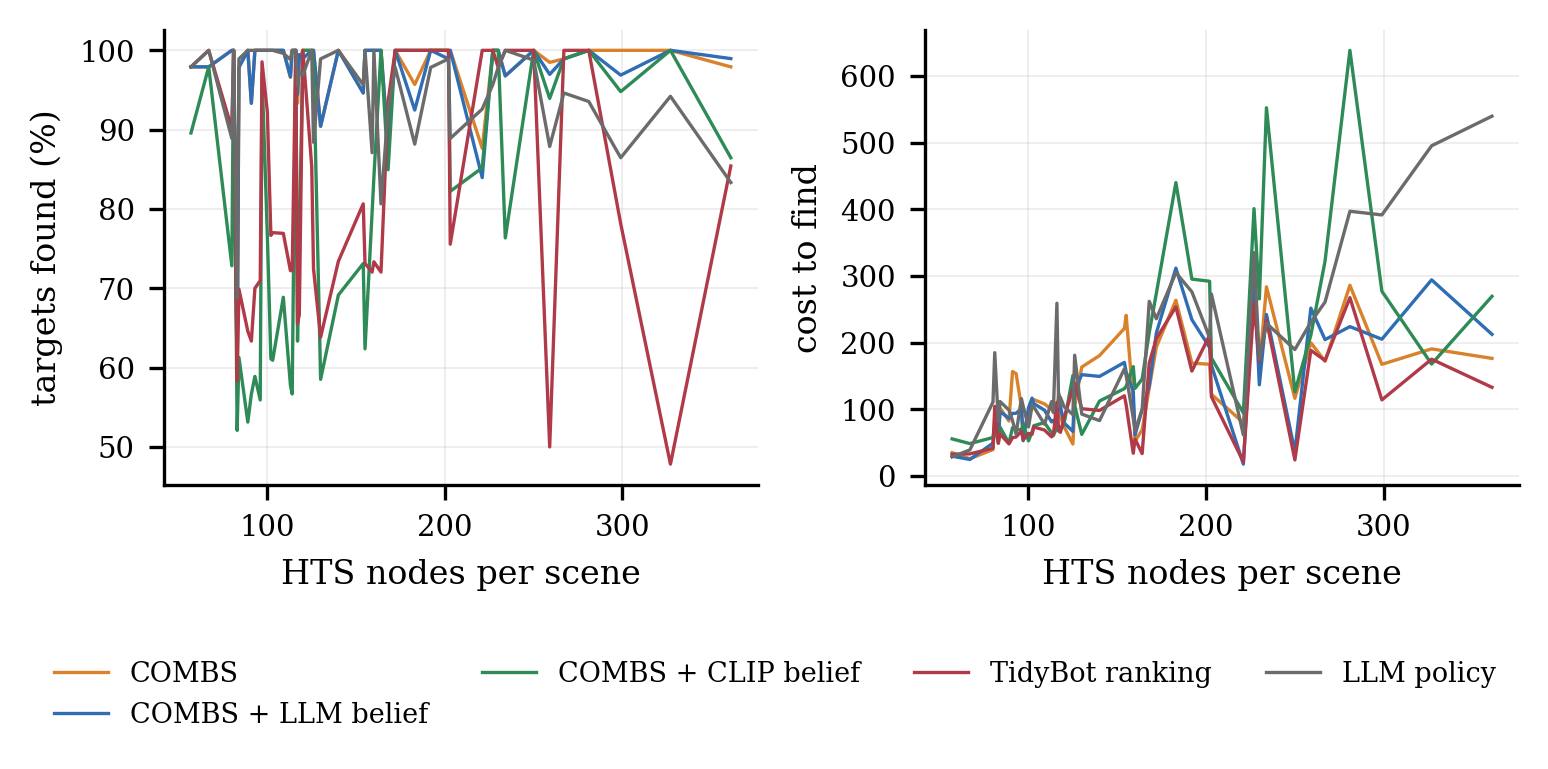}
\caption{Success rate and mean cost of found targets against scene size, measured as the largest HTS any episode revealed in the scene, one point per scene, pooled over the simulators.
}
\label{fig:hts_size}
\end{figure}

\begin{figure}
\centering
\includegraphics[width=\columnwidth]{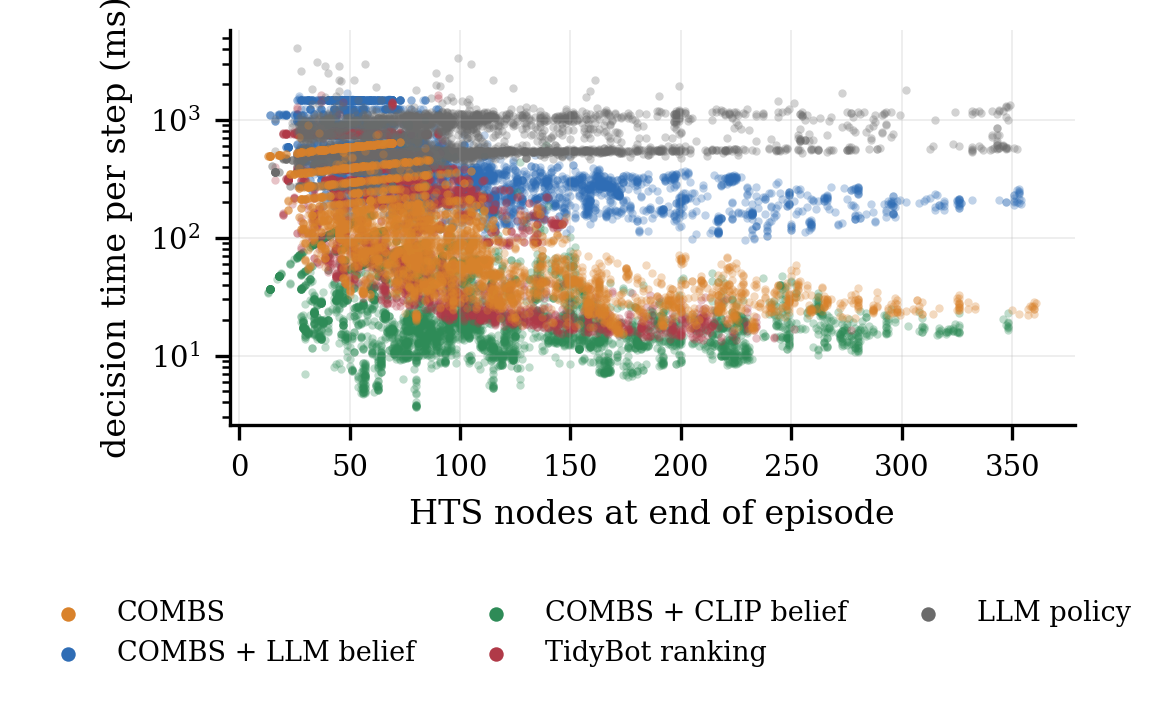}
\caption{Decision time per search step against the number of HTS nodes at the end of each episode, for each method, pooled over simulators, on a log scale. Each episode is charged its external calls at their measured latencies. 
}
\label{fig:compute_methods}
\end{figure}

COMBS median decision time is 1.7--3.1~ms on every simulation. Success does not fall as the HTS grows, except mildly on MP3D Detic. \Cref{fig:compute_methods} compares planning time across methods: COMBS falls in the same range as COMBS + CLIP belief and the TidyBot ranking, and well below the external LLM methods, at every graph size, so its concealment reasoning and absence test add no noticeable planning cost.
Most of the HTS size is attributed to nodes revealed during search, and COMBS's success holds across graph sizes, falling only where perception is noisy and the graph is large.

Considering decision latency from model calls, 
COMBS spends 36--96~ms, COMBS + LLM belief 210--366~ms, the per-step LLM policy 489--563~ms, and Tidybot ranking 33--262~ms, so the COMBS prior supplies knowledge at a fraction of an external LLM's per-step latency.

\paragraph{Scale}

\Cref{fig:hts_size} relates performance to how much there is to search. COMBS and COMBS + LLM belief find nearly every target at every scene size, while the TidyBot ranking and COMBS + CLIP belief miss targets on many individual scenes; cost rises with HTS size for every method, with the largest spikes for COMBS + CLIP belief.


\section{Conclusion}
We present COMBS and the CA-MOMDP, to our knowledge the first Bayes-adaptive mixed-observability formulation for finding a target object that may be concealed, combining a zero-shot prior fused from three semantic sources with bounded influence for each, an ontology-derived hypothesis over what may conceal, and a negative-search update that never lets a look rule out a closed container. 
Future work includes reasoning between same-room same-class instances based on their 
spatial context, since nodes currently share their room's belief which can dilute prior values.

\bibliographystyle{ieeetr}
\bibliography{main.bib}
\end{document}